\documentclass{bmvc2k}
\usepackage{diagbox}

\title{A Deep Learning Perspective on the Origin of Facial Expressions}

\addauthor{Ran Breuer}{rbreuer@cs.technion.ac.il}{1}
\addauthor{Ron Kimmel}{ron@cs.technion.ac.il}{1}

\addinstitution{
Department of Computer Science\\
Technion - Israel Institute of Technology\\
Technion City, Haifa, Israel
}

\runninghead{Breuer, Kimmel}{A Deep Learning Perspective on Facial Expressions}

\def\etal{\emph{et al}\bmvaOneDot}

\begin{document}

\maketitle

\begin{figure*}[th]
\begin{center}
	\includegraphics[width=\textwidth]{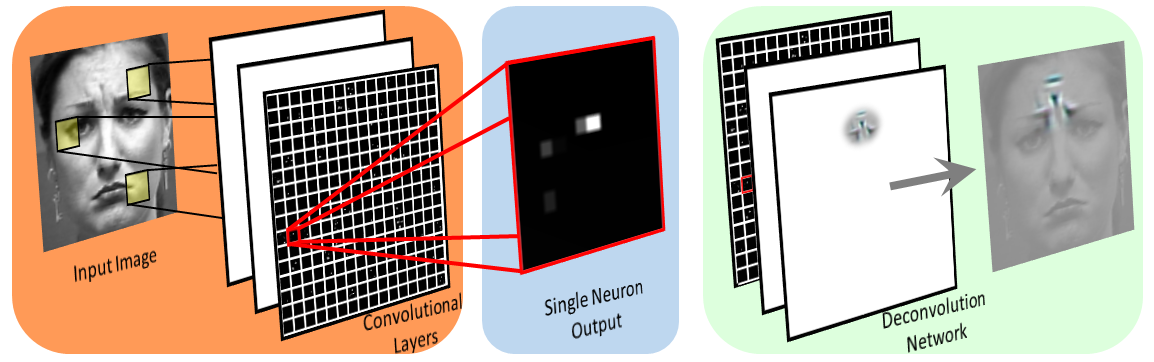}
\end{center}
\caption{\label{fig:gbp_process} Demonstration of the filter visualization process.}
\end{figure*}

\begin{abstract}
Facial expressions play a significant role in human communication and behavior.
Psychologists have long studied the relationship between facial expressions and emotions. 
Paul Ekman \etal \cite{ekman70,ekman78FACS}, devised the Facial Action Coding System (FACS) 
 to taxonomize human facial expressions and model their behavior.
The ability to recognize facial expressions automatically, enables novel applications  
 in fields like human-computer interaction, social gaming, and psychological research.
There has been a tremendously active research in this field, with several recent papers
 utilizing convolutional neural networks (CNN) for feature extraction and inference.
In this paper, we employ CNN understanding methods to study the relation between the 
 features these computational networks are using, the FACS and Action Units (AU).
We verify our findings on the Extended Cohn-Kanade (CK+), NovaEmotions and FER2013 datasets.
We apply these models to various tasks and tests using transfer learning,
 including cross-dataset validation and cross-task performance.
Finally, we exploit the nature of the FER based CNN models for the detection of micro-expressions 
 and achieve state-of-the-art accuracy using a simple long-short-term-memory (LSTM) 
  recurrent neural network (RNN).
\end{abstract}

\section{Introduction} \label{Intro}

Human communication consists of much more than verbal elements, words and sentences. 
Facial expressions (FE) play a significant role in inter-person interaction. 
They convey emotional state, truthfulness and add context to the verbal channel.
Automatic FE recognition (AFER) is an interdisciplinary domain standing at the crossing
 of behavioral science, psychology, neurology, and artificial intelligence.
 
\smallskip
\subsection{Facial Expression Analysis}
\label{Facial Expressions Analysis}
\smallskip

\noindent The analysis of human emotions through facial expressions is a major part in psychological research. 
Darwin's work in the late 1800's \cite{DarwinFE} placed human facial expressions within an
 evolutionary context. 
Darwin suggested that facial expressions are the residual actions of more complete behavioral
 responses to environmental challenges. 
When in disgust, constricting the nostrils served to reduce inhalation of noxious or
 harmful substances. 
Widening of the eyes in surprise increased the visual field to better see an 
 unexpected stimulus.

\begin{figure}[t]
\begin{center}
   \includegraphics[width=0.8\linewidth]{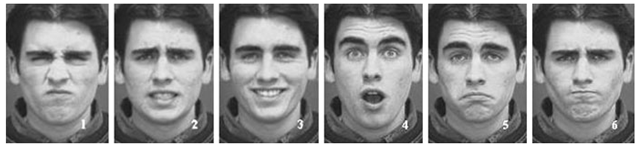}
\end{center}
   \caption{Example of primary universal emotions. 
      From left to right: 
     disgust, fear, happiness, surprise, sadness, and anger.\protect\footnotemark}
\label{fig:expressions}
\end{figure}

Inspired by Darwin's evolutionary basis for expressions, Ekman \etal \cite{ekman70} introduced 
 their seminal study about facial expressions. 
They identified seven primary, universal expressions where universality related to the 
 fact that these expressions remain the same across different cultures \cite{ekman94}. 
Ekman labeled them by their corresponding emotional states, that is,  
  \emph{happiness, sadness, surprise, fear, disgust, anger}, and \emph{contempt}
  , see Figure \ref{fig:expressions}. 
Due to its simplicity and claim for universality, 
 the primary emotions hypothesis has been extensively exploited in
 cognitive computing.

\begin{figure}[htb]
\begin{center}
	\includegraphics[width=\linewidth]{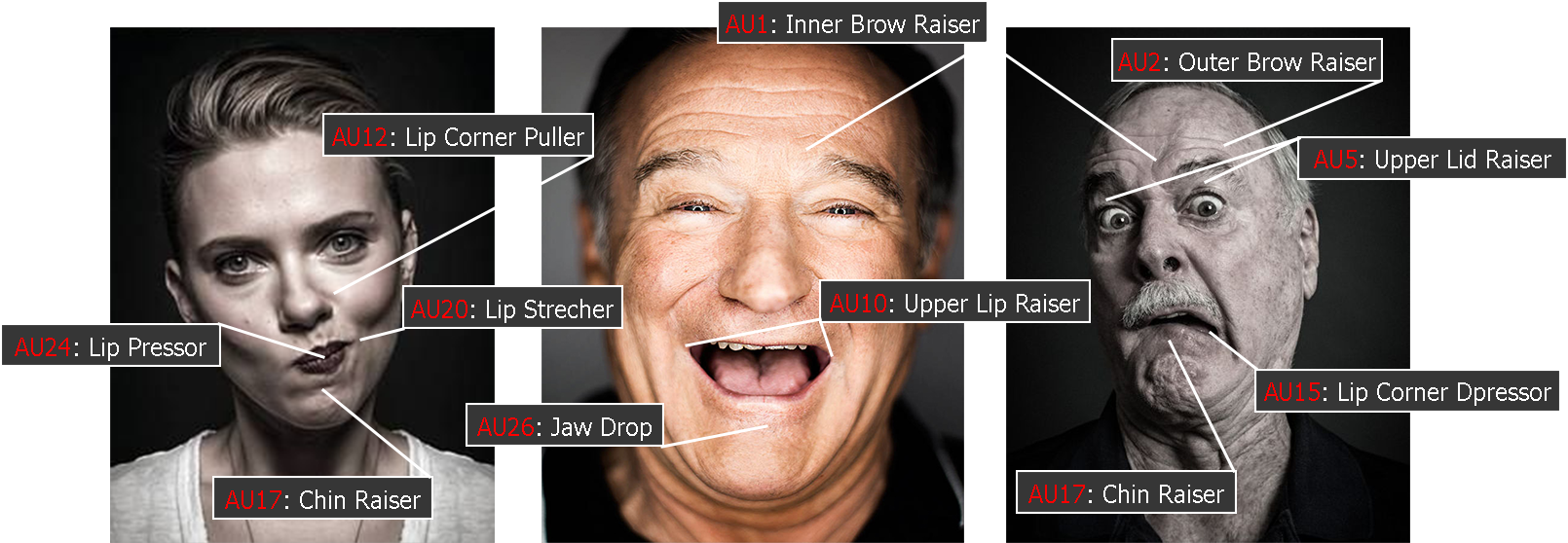}
\end{center}
\caption{\label{fig:facs_demo}
 Expressive images and their active AU coding. 
 This demonstrates the composition of describing one's facial expression using a collection
 of FACS based descriptors.
 }
\end{figure}

\footnotetext{Images taken from \cite{schmidt2001human} \copyright Jeffrey Cohn}
In order to further investigate emotions and their corresponding facial expressions, 
 Ekman devised the \emph{facial action coding system} (FACS) \cite{ekman78FACS}. 
FACS is an anatomically based system for describing all observable facial movements
 for each emotion, see Figure \ref{fig:facs_demo}. 
Using FACS as a methodological measuring system, one can describe any expression
 by the \emph{action units} (AU) one activates and its activation intensity.
Each action unit describes a cluster of facial muscles that act together
 to form a specific movement.
According to Ekman, there are $44$ facial AUs, describing actions such as ``open mouth'', 
 ``squint eyes'' etc., and $20$ other AUs were added in a $2002$ revision of the 
  FACS manual \cite{ekman2002}, to account for head and eye movement.

\smallskip
\subsection{Facial Expression Recognition and Analysis} \label{FER}
\smallskip

\noindent The ability to automatically recognize facial expressions and infer the emotional state
 has a wide range of applications. 
These included emotionally and socially aware systems  
 \cite{integrating2002,vinciarelli2009social,devault2014simsensei}, improved gaming 
  experience \cite{bakkes2012personalised}, driver drowsiness detection \cite{vural2007drowsy},
  and detecting pain in patients \cite{lucey2011automaticallypain} 
  as well as distress \cite{joshi2012neuraldepress}. 
Recent papers have even integrated automatic analysis of viewers' reaction for the  
 effectiveness of advertisements \cite{emotient,affectiva,realeyes}.

Various methods have been used for \emph{automatic facial expression recognition} (FER or AFER) tasks.
Early papers used geometric representations, for example, vectors descriptors
 for the motion of the face \cite{motionvectors}, active contours for mouth and 
 eye shape retrieval \cite{activecontourFER}, and using 2D deformable mesh models \cite{kotsia2007aam}.
Other used appearance representation based methods, such as Gabor filters \cite{gaborFER},
 or local binary patterns (LBP) \cite{shan2009facialLBP}. 
These feature extraction methods usually were combined with one of several regressors to translate
 these feature vectors to emotion classification or action unit detection. 
The most popular regressors used in this context were support vector machines (SVM) and random forests. 
For further reading on the methods used in FER, we refer the reader to  
 \cite{martinez2016advances,SurveyFACSDetect2016,Survey09,FACS-ITW-Survey}

\smallskip
\subsection{Understanding Convolutional Neural Networks} \label{CNN}
\smallskip

\noindent Over the last part of this past decade, \emph{convolutional neural networks} (CNN) \cite{CNNLeCun} 
 and \emph{deep belief networks} (DBN) have been used for feature extraction, classification
 and recognition tasks. 
These CNNs have achieved state-of-the-art results in various fields, including object recognition
 \cite{ImageNet12}, face recognition \cite{DeepFaceRecog}, and scene understanding \cite{zhou2014learning}. Leading challenges in FER \cite{FERA2013,FERA2015,valstar2016avec} have also been led by methods 
  using CNNs \cite{CNN-FACS-Gudi15,LeCunCNNFaceApplication,ghosh2015multi}.

Convolutional neural networks, as first proposed by LeCun in 1998 \cite{CNNLeCun}, employ concepts 
 of receptive fields and weight sharing. 
The number of trainable parameters is greatly reduced and the propagation of information through
 the layers of the network can be simply calculated by convolution. 
The input, like an image or a signal, is convolved through a filter collection (or map) in the
 \emph{convolution layers} to produce a feature map. 
Each feature map detects the presence of a single feature at all possible input locations.

In the effort of improving CNN performance, researchers have developed methods of exploring and
 understanding the models learned by these methods. 
\cite{DeepInsideCNN13} demonstrated how saliency maps can be obtained from a ConvNet by projecting 
 back from the fully connected layers of the network. 
\cite{girshick2014rich} showed visualizations that identify patches within a dataset that are 
 responsible for strong activations at higher layers in the model. 

Zeiler \etal \cite{Zeiler10deconvolutionalnetworks,Zeiler2014DeconvViz} describe using 
 \emph{deconvolutional networks} as a way to visualize a single unit in a feature map of a given CNN, 
  trained on the same data. 
The main idea is to visualize the input pixels that cause a certain neuron, like a filter 
 from a convolutional layer, to maximize its output. 
This process involves a \emph{feed forward} step, where we stream the input through the network,
 while recording the consequent activations in the middle layers. 
Afterwards, one fixes the desired filter's (or neuron) output, and sets all other elements to 
 the neutral elements (usually 0). 
Then, one ``back-propagates'' through the network all the way to the input layer, where we would 
 get a neutral image with only a few pixels set - those are the pixels responsible for max activation
 in the fixed neuron. 
Zeiler \etal. found that while the first layers in the CNN model seemed to learn Gabor-like filters,
 the deeper layers were learning high level representations of the objects the network was trained
 to recognize.
By finding the maximal activation for each neuron, and back-propagating through the deconvolution layers,
 one could actually view the locations that caused a specific neuron to react.

Further efforts to understand the features in the CNN model, were done by Springenberg \etal. 
 who devised \emph{guided back-propagation} \cite{GuidedBP14}. 
With some minor modifications to the deconvolutional network approach, they were able to produce
 more understandable outputs, which provided better insight into the model's behavior.
The ability to visualize filter maps in CNNs improved the capability of understanding what 
 the network learns during the training stage.

\smallskip

\noindent The main contributions of this paper are as follows.
\begin{itemize}

\item  We employ CNN visualization techniques to understand the model learned by current 
 state-of-the-art methods in FER on various datasets. 
 We provide a computational justification for Ekman's FACS\cite{ekman78FACS} as a leading model 
 in the study of human facial expressions.

\item We show the generalization capability of networks trained on emotion detection, 
 both across datasets and across various FER related tasks.

\item We discuss various applications of FACS based feature representation produced by 
 CNN-based FER methods.
\end{itemize}

\section{Experiments} \label{Experiments}

Our goal is to explore the knowledge (or models) as learned by state-of-the-art methods for FER, similar to the works of \cite{khorrami2015deep}. 
We use CNN-based methods on various datasets to get a sense of a common model structure, 
 and study the relation of these models to Ekman's FACS \cite{ekman78FACS}.
To inspect the learned models ability to generalize, we use the method of \emph{transfer learning}
 \cite{Transfer14Yosinski} to see how these models perform on other datasets.
We also measure the models' ability to perform on other FER related tasks, 
 ones which they were not explicitly trained for.

\label{Datasets}

\begin{figure}[htb]
\begin{center}
  \begin{tabular}{@{}c@{}}
    \includegraphics[width=.125\linewidth]{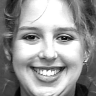}
    \includegraphics[width=.125\linewidth]{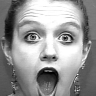}
    \includegraphics[width=.125\linewidth]{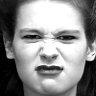}
    \includegraphics[width=.125\linewidth]{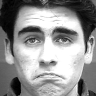}
    \includegraphics[width=.125\linewidth]{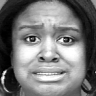}
    \includegraphics[width=.125\linewidth]{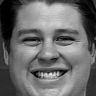}
  \end{tabular}
  
  \begin{tabular}{@{}c@{}}
  	\includegraphics[width=.75\linewidth]{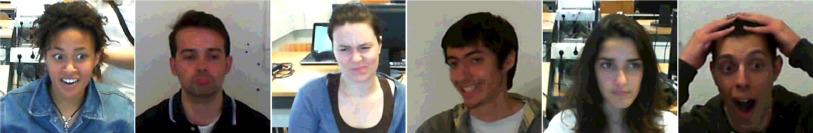}
  \end{tabular}
  
  \begin{tabular}{@{}c@{}}
    \includegraphics[width=.125\linewidth]{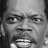}
    \includegraphics[width=.125\linewidth]{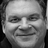}
    \includegraphics[width=.125\linewidth]{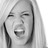}
    \includegraphics[width=.125\linewidth]{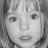}
    \includegraphics[width=.125\linewidth]{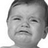}
    \includegraphics[width=.125\linewidth]{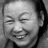}
  \end{tabular}
\end{center}
\caption{\label{fig:dataset_example}Images from CK+ (top), NovaEmotions (middle) and FER2013 (bottom) datasets.}
\end{figure}

In order to get a sense of the common properties of CNN-based state-of-the-art models in FER, 
 we employ these methods on numerous datasets. 
Below are brief descriptions of datasets used in our experiments. 
See Figure \ref{fig:dataset_example} for examples.

\smallskip
{\bf{Extended Cohn-Kanade}}
\smallskip
The \emph{Extended Cohn-Kanade} dataset (CK+) \cite{CKplus}, is comprised of video sequences
 describing the facial behavior of 210 adults. 
Participant ages range from 18 to 50. 69\% are female, 91\% Euro-American, 13\% Afro-American,
 and 6\% belong to other groups. 
The dataset is composed of 593 sequences from 123 subjects containing posed facial expressions. 
Another 107 sequences were added after the initial dataset was released.
These sequences captured spontaneous expressions performed between formal sessions
 during the initial recordings, that is, non-posed facial expressions.

Data from the Cohn-Kanade dataset is labeled for emotional classes (of the 7 primary emotions
 by Ekman \cite{ekman70}) at peak frames. 
In addition, AU labeling was done by two certified FACS coders. 
Inter-coder agreement verification was performed for all released data.

\smallskip
{\bf{NovaEmotions}}
\smallskip
\emph{NovaEmotions} \cite{NovaEmo1,NovaEmo2}, aim to represent facial expressions and emotional state
 as captured in a non-controlled environment. 
The data is collected in a crowd-sourcing manner, where subjects were put in front of a gaming device, 
 which captured their response to scenes and challenges in the game itself. 
The game, in time, reacted to the player's response as well. 
This allowed collecting spontaneous expressions from a large pool of variations.

The NovaEmotions dataset consists of over 42,000 images taken from 40 different people. 
Majority of the participants were college students with ages ranges between 18 and 25. 
Data presents a variety of poses and illumination. 
In this paper we use cropped images containing only the face regions. 
Images were aligned such that eyes are presented on the same horizontal line across 
 all images in the dataset. 
Each frame was annotated by multiple sources, both certified professionals as
 well as random individuals. 
A consensus was collected for the annotation of the frames, resulting in the final labeling.

\smallskip
{\bf{FER 2013}}
\smallskip
The \emph{FER 2013} challenge \cite{FER2013} was created using Google image search API with 184
 emotion related keywords, like \emph{blissful, enraged}. 
Keywords were combined with phrases for gender, age and ethnicity in order to obtain up to 600 
 different search queries. 
Image data was collected for the first 1000 images for each query. 
Collected images were passed through post-processing, that involved
 face region cropping and image alignment. 
Images were then grouped into the corresponding fine-grained emotion classes, rejecting wrongfully 
 labeled frames and adjusting cropped regions. 
The resulting data contains nearly 36,000 images, divided into 8 classes (7 effective expressions 
 and a neutral class), with each emotion class containing a few thousand images (\emph{disgust}
 being the exception with only 547 frames).

\medskip
\subsection{Network Architecture and Training}
\smallskip

\noindent For all experiments described in this paper, we implemented a simple, classic
 \emph{feed-forward convolutional neural network}. 
Each network is structured as follows. 
An input layer, receiving a gray-level or RGB image. 
The input is passed through 3 convolutional layer blocks, each block consists of a filter map layer,
 a non-linearity (or activation) and a max pooling layer. 
Our implementation is comprised of 3 convolutional blocks, each with a \emph{rectified linear unit} 
 (ReLU \cite{ReluDropout}) activation and a pooling layer with $2\times2$ pool size. 
The convolutional layers have filter maps with increasing filter (neuron) count the deeper the 
 layer is, resulting in a 64, 128 and 256 filter map sizes, respectively. 
Each filter in our experiments supports $5\times5$ pixels.

The convolutional blocks are followed by a fully-connected layer with 512 hidden neurons. 
The hidden layer's output is transferred to the output layer, which size is affected by the 
 task in hand, 8 for emotion classification, and up to 50 for AU labeling. 
The output layer can vary in activation, for example, for classification tasks we 
 prefer \emph{softmax}.

To reduce over-fitting, we used \emph{dropout} \cite{ReluDropout}. 
We apply the dropout after the last convolutional layer and between the fully-connected layers, 
 with probabilities of 0.25 and 0.5 respectively. 
A dropout probability $p$ means that each neuron's output is set to 0 with probability $p$.

We trained our network using ADAM \cite{ADAMOptimizer} optimizer with a learning rate of
 $1e-3$ and a decay rate of $1e-5$.
To maximize generalization of the model, we use methods of data augmentation.
We use combinations of random flips and affine transforms, e.g. rotation, translation, scaling, 
 sheer, on the images to generate synthetic data and enlarge the training set.
Our implementation is based on the Keras \cite{keras} library with \emph{TensorFlow}
 \cite{tensorflow} back-end. 
We use OpenCV \cite{opencv} for all image operations.

\section{Results and Analysis} 
\label{results}
We verify the performance of our networks on the datasets mentioned in \ref{Datasets} using a 
 10-fold cross validation technique. 
For comparison, we use the frameworks of \cite{AUDN,shan2009facialLBP,FER2013,gaborFER}. 
We analyze the networks' ability to classify facial expression images into the 7 primary emotions
 or as a neutral pose. 
Accuracy is measured as the average score of the 10-fold cross validation. 
Our model performs at state-of-the-art level when compared to the leading methods in AFER,
See Tables \ref{tab:acc_ck},\ref{tab:acc_fer13}.

\begin{table}
\parbox{.4\linewidth}{
\centering
\begin{tabular}{|l|c|}
	\hline
    Method & Accuracy \\
    \hline\hline
    Gabor+SVM \cite{gaborFER} & 89.8\% \\
    LBPSVM \cite{shan2009facialLBP} & 95.1\% \\
    AUDN \cite{AUDN} & 93.70\% \\
    BDBN \cite{BDBNFacial} & 96.7\% \\
    \textbf{Ours} & 98.62 \% $\pm$ 0.11\% \\
    \hline
\end{tabular}
\caption{\label{tab:acc_ck}
Accuracy evaluation of emotion classification on the CK+ dataset.
}
}
\hfill
\parbox{.4\linewidth}{
\centering
\begin{tabular}{|l|c|}
	\hline
    Method & Accuracy \\
    \hline\hline
    Human Accuracy & 68\% $\pm$ 5\% \\
    RBM & 71.162\% \\
    VGG CNN \cite{VGG} & 72.7\% \\
    ResNet CNN \cite{ResNet} & 72.4\% \\
    \textbf{Ours} & 72.1\% $\pm$ 0.5\% \\
    \hline
\end{tabular}
\caption{\label{tab:acc_fer13}
 Accuracy evaluation of emotion classification on the FER 2013 challenge. 
 Methods and scores are documented in \cite{FER2013,martinez2016advances}.
 }
 }
\end{table}

\smallskip
\subsection{Visualizing the CNN Filters}
\smallskip

\noindent After establishing a sound classification framework for emotions, we move to analyze the 
 models that were learned by the suggested network.
We employ Zeiler \etal and Springenberg's \cite{Zeiler2014DeconvViz,GuidedBP14} methods for 
 visualizing the filters trained by the proposed networks on the different emotion classification
  tasks, see Figure \ref{fig:gbp_process}. 

\begin{figure*}[htb]
\begin{center}
  \begin{tabular}{@{}c@{}}
    \includegraphics[width=.125\linewidth]{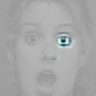}
    \includegraphics[width=.125\linewidth]{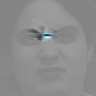}
    \includegraphics[width=.125\linewidth]{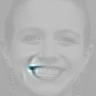}
    \includegraphics[width=.125\linewidth]{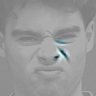}
    \includegraphics[width=.125\linewidth]{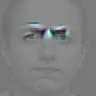}
    \includegraphics[width=.125\linewidth]{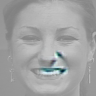}
    \includegraphics[width=.125\linewidth]{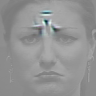}
    \\[\abovecaptionskip]
  \end{tabular}

  \begin{tabular}{@{}c@{}}
    \includegraphics[width=.125\linewidth]{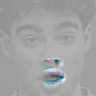}
    \includegraphics[width=.125\linewidth]{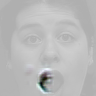}
    \includegraphics[width=.125\linewidth]{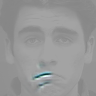}
    \includegraphics[width=.125\linewidth]{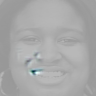}
    \includegraphics[width=.125\linewidth]{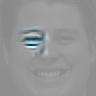}
    \includegraphics[width=.125\linewidth]{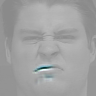}
    \includegraphics[width=.125\linewidth]{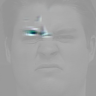}
    \\[\abovecaptionskip]
  \end{tabular}
\end{center}
\caption{\label{fig:feat_viz1}
	Feature visualization for the trained network model. 
   For each feature we overlay the deconvolution output on top of its original input image.
   One can easily see the regions to which each feature refers.
    }
\end{figure*}

\begin{figure*}[htb]
\begin{center}
\begin{tabular}{@{}c@{}}
  \begin{tabular}{@{}c@{}}
    \includegraphics[width=.15\linewidth]{images/overlay9}
    \includegraphics[width=.1\linewidth]{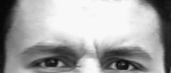}			
    \\
    \small{AU4: Brow lowerer}
  \end{tabular}
  
  \begin{tabular}{@{}c@{}}
    \includegraphics[width=.15\linewidth]{images/overlay1}
    \includegraphics[width=.1\linewidth]{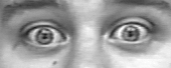}			
    \\
    \small{AU5: Upper lid raiser}
  \end{tabular}
  
  \begin{tabular}{@{}c@{}}
    \includegraphics[width=.15\linewidth]{images/overlay17}
    \includegraphics[width=.1\linewidth]{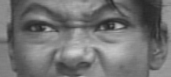}			
    \\
    \small{AU9: Nose wrinkler}
  \end{tabular}
\end{tabular}
\begin{tabular}{@{}c@{}}
  \begin{tabular}{@{}c@{}}
    \includegraphics[width=.15\linewidth]{images/overlay5}
    \includegraphics[width=.1\linewidth]{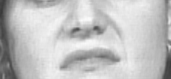}			
    \\
    \small{AU10: Upper lip raiser}
  \end{tabular}

  \begin{tabular}{@{}c@{}}
    \includegraphics[width=.15\linewidth]{images/overlay3}
    \includegraphics[width=.1\linewidth]{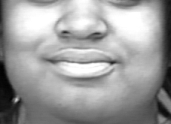}			
    \\
    \small{AU12: Lip Corner Puller}
  \end{tabular}
  
  \begin{tabular}{@{}c@{}}
    \includegraphics[width=.15\linewidth]{images/overlay6}
    \includegraphics[width=.1\linewidth]{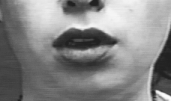}			
    \\
    \small{AU25: Lips Part}
  \end{tabular}
\end{tabular}
\end{center}
\caption{\label{fig:feat_viz_facs}Several feature maps and their corresponding FACS Action Unit.}
\end{figure*}

As shown by \cite{Zeiler2014DeconvViz}, the lower layers provide low level Gabor-like filters whereas 
 the mid and higher layers, that are closer to the output, provide high level, human readable features. 
By using the methods above, we visualize the features of the trained network. 
Feature visualization is shown in Figure \ref{fig:feat_viz1} through input that maximized activation 
 of the desired filter alongside the pixels that are responsible for the said response. 
From analyzing the trained models, one can notice great similarity between our networks' feature 
 maps and specific facial regions and motions. 
Further investigation shows that these regions and motions have significant correlation to
 those used by Ekman to define the FACS Action Units, see Figure \ref{fig:feat_viz_facs}.

We matched a filter's suspected AU representation with the actual CK+ AU labeling, using the following method.
\begin{enumerate}
\item Given a convolutional layer $l$ and filter $j$, the activation output is marked as $F_{l,j}$.
\item We extracted the top $N$ input images that maximized, $i =  \arg_i\max F_{l,j}(i)$. 
\item For each input $i$, the manually annotated AU labeling is $A_i^{44\times 1}$. 
     $A_{i,u}$ is 1 if AU $u$ is present in $i$.
\item The correlation of filter $j$ with AU $u$'s presence is $P_{j,u}$ and is defined by
$P_{j,u}=\frac{\sum A_{i,u}}{N}.$ 
\end{enumerate}

Since we used a small $N$, we rejected correlations with $P_{j,u}<1$. 
Out of 50 active neurons from a 256 filters map trained on CK+, only 7 were rejected. 
This shows an amazingly high correlation between a CNN-based model, trained with no prior knowledge, and Ekman's facial action coding system (FACS).


In addition, we found that even though some AU-inspired filters were created more than just once,
 a large amount of neurons in the highest layers were found ``dead'', that is,
 they were not producing effective output for any input.
The amount of active neurons in the last convolutional layer was about 30\%
 of the feature map size ($60$ out of $256$).
The number of effective neurons is similar to the size of Ekman's vocabulary
 of action units by which facial expressions can be identified.

\smallskip
\subsection{Model Generality and Transfer Learning}\label{TransferLearn}
\smallskip

\noindent After computationally demonstrating the strong correlation between Ekman's FACS and the
 model learned by the proposed computational neural network, we study the model's 
 ability to generalize and solve other problems related to expression recognition
 on various data sets. 
We use the \emph{transfer learning} training methodology \cite{Transfer14Yosinski} 
 and apply it to different tasks. 

Transfer learning, or \emph{knowledge transfer}, aims to use models that were pre-trained  
 on different data for new tasks.
Neural network models often require large training sets.
However, in some scenarios the size of the training set is insufficient for proper training. 
Transfer learning allows using the convolutional layers as pre-trained feature extractors, 
 with only the output layers being replaced or modified according to the task at hand.
That is, the first layers are treated as pre-defined features,  while the last 
 layers, that define the task at hand, are adapted by learning based on the available training set.

We tested our models on both cross-dataset and cross-task capabilities.
In most FER related tasks, AU detection is done as a leave-one-out manner. 
Given an input (image or video) the system would predict the probability of a specific AU to be active. 
This method is proven to be more accurate than training against the detection of all AU activations
 at the same time, mostly due to the sizes of the training datasets.
When testing our models against detection of a single AU, we recorded high accuracy scores with most AUs.
Some action units, like AU11: \emph{nasolabial deepener}, were not predicted properly 
 in some cases when using the suggested  model. 
 A better prediction model for these AUs would require a dedicated set of features that focus on the relevant region in the face, since they signify a minor facial movement.


The leave-one-out approach is commonly used since the training set is not large enough to 
 train a classifier for all AUs simultaneously (all-against-all). 
In our case, predicting all AU activations simultaneously for a single image, requires a larger
 dataset than the one we used. 
Having trained our model to predict only eight classes, we verify our model on an 
 all-against-all approach and obtained result that compete with the leave-one-out classifiers.
In order to increase accuracy, we apply a sparsity inducing loss function on the output layer by combining both $L_2$ and $L_1$ terms. 
This resulted in a sparse FACS coding of the input frame. 
When testing for binary representation, that is, only an active/nonactive prediction per AU, 
 we recorded an accuracy rate of 97.54\%. 
When predicting AU intensity, an integer of range 0 to 5, we recorded an accuracy rate of 96.1\% 
 with a mean square error (MSE) of 0.2045.

When testing emotion detection capabilities across datasets, we found that the trained models had
 very high scores. 
This shows, once again, that the FACS-like features trained on one dataset can be applied almost 
 directly to another, see Table \ref{tab:cross_ds}. 

\begin{table}
\begin{center}
\begin{tabular}{|c|c|c|c|}
\hline
\backslashbox{Train}{Test} & CK+ & FER2013 & NovaEmotions \\
\hline
CK+ & 98.62\% & 69.3\% &  67.2\% \\
\hline
FER2013 & 92.0\% & 72.1\% &  78.0\% \\
\hline
NovaEmotions & 93.75\% & 71.8\% & 81.3\% \\
\hline
\end{tabular}
\end{center}
\caption{\label{tab:cross_ds} Cross dataset application of emotion detection models.}
\end{table}

\section{Micro-Expression Detection} \label{MEDetect}

\emph{Micro-expressions} (ME) are a more spontaneous and subtle facial movements that happen involuntarily, 
 thus reveling one's genuine, underlying emotion \cite{ekman1969nonverbal}. 
These micro-expressions are comprised of the same facial movements that define FACS action units and 
 differ in intensity. 
ME tend to last up to 0.5sec, making detection a challenging task for an un-trained individual. 
Each ME is broken down to 3 steps: Onset, apex, and offset, describing the beginning, peek, and 
 the end of the motion, respectively. 

Similar to AFER, a significant effort was invested in the last years to train computers in order 
 to automatically detect micro-expressions and emotions. 
Due to its low movement intensity, automatic detection of micro-expressions requires a temporal sequence,
 as opposed to a single frame. 
Moreover, since micro-expressions tend to last for just a short time and occur in a brief of a moment, 
 a high speed camera is usually used for capturing the frames. 

We apply our FACS-like feature extractors to the task of automatically detecting micro-expressions. 
To that end, we use the \emph{CASME II} dataset \cite{CASME2}.
CASME II includes 256 spontaneous micro-expressions filmed at 200fps. 
All videos are tagged for \emph{onset, apex, and offset} times, as well as the expression conveyed. 
AU coding was added for the \emph{apex} frame. 
Expressions were captured by showing a subject video segments that triggered the desired response.

To implement our micro-expressions detection network, we first trained the network on selected frames from the training
 data sequences. 
For each video, we took only the \emph{onset, apex, and offset} frames, as well as the first and last 
 frames of the sequence, to account for neutral poses. 
Similar to Section \ref{TransferLearn}, we first trained our CNN to detect emotions.
We then combined the convolutional layers from the trained network, 
 with a \emph{long-short-tern-memory} \cite{LSTM} recurrent neural network (RNN),
 whose input is connected to the first fully connected layer of the feature extractor CNN. 
The LSTM we used is a very shallow network, with only a LSTM layer and an output layer. 
Recurrent dropout was used after the LSTM layer. 

We tested our network with a leave-one-out strategy, where one subject was designated as test 
 and was left out of training.
Our method performs at state-of-the-art level (Table \ref{tab:meperf}).

\begin{table}
\begin{center}
\begin{tabular}{|l|c|}
\hline
Method & Accuracy \\
\hline\hline
LBP-TOP \cite{LBPTOP2007} & 44.12\% \\
LBP-TOP with adaptive magnification \cite{subtleLBPMagnification} & 51.91\% \\
\textbf{Ours} & 59.47\% \\
\hline
\end{tabular}
\end{center}
\caption{\label{tab:meperf}	
Micro-expression detection and analysis accuracy.
Comparison with reported state-of-the-art methods.
}
\end{table}

\section{Conclusions}
We provided a computational justification of Ekman's facial action 
 units (FACS) which is the core of his facial expression analysis axiomatic/observational
 framework.
We studied the models learned by state-of-the-art CNNs, 
 and used CNN visualization techniques to understand the feature maps that are obtained 
 by training for emotion detection of seven universal expressions.
We demonstrated a strong correlation between the features generated by an unsupervised
 learning process and Ekman's action units used as the atoms in his leading facial
 expressions analysis methods.
The FACS-based features' ability to generalize was then verified on cross-data and cross-task 
 aspects that provided high accuracy scores.
Equipped with refined computationally-learned action units that align with Ekman's theory, 
 we applied our models to the task of micro-expression detection and obtained recognition rates that outperformed state-of-the-art methods.

The FACS based models can be further applied to other FER related tasks. 
Embedding emotion or micro-expression recognition and analysis as part of real-time applications
 can be useful in several fields, for example, lie detection, gaming, and marketing analysis.
Analyzing computer generated recognition models can help refine Ekman's theory of reading 
 facial expressions and emotions and provide an even better support for its validity and accuracy.


\bibliography{general}
\end{document}